\documentclass[journal]{IEEEtran}
\usepackage[utf8]{inputenc}
\usepackage{multirow}
\IEEEoverridecommandlockouts
\usepackage{cite}
\usepackage{tabularx}
\usepackage{amsfonts}
\usepackage{amsmath}
\usepackage{amssymb}
\usepackage{array}
\usepackage{bm}
\usepackage{cases}
\usepackage{color}
\usepackage{enumitem}
\usepackage[dvipsnames]{xcolor} 
 
\usepackage[overload]{empheq}
\usepackage{graphicx}   
\usepackage{makecell}
\usepackage{mathrsfs}
\usepackage{multirow} 
\usepackage{threeparttable} 

\usepackage{physics}    
\usepackage{soul}

\usepackage{subcaption}
\usepackage{xspace}

\usepackage{url}
\usepackage[all]{xy}
\usepackage[font=footnotesize,labelsep=period]{caption}

\newcommand{\PreserveBackslash}[1]{\let\temp=\\#1\let\\=\temp}
\newcolumntype{C}[1]{>{\PreserveBackslash\centering}p{#1}}
\newcolumntype{R}[1]{>{\PreserveBackslash\raggedleft}p{#1}}
\newcolumntype{L}[1]{>{\PreserveBackslash\raggedright}p{#1}}

\usepackage{tikz}   
\usepackage{standalone}
\usepackage{pgfplots}
\usetikzlibrary{shapes, arrows,chains,calc,shapes}
\pgfplotsset{compat = newest}
\pgfdeclarelayer{bg}    
\pgfsetlayers{bg,main}  

\usepackage{amsmath,amssymb,amsfonts}
\usepackage{graphicx}
\usepackage{textcomp}
\usepackage{amssymb}
\usepackage{xcolor}
\usepackage{xspace}
\usepackage{comment}
\usepackage{algorithm2e}
\usepackage{algpseudocode}
\graphicspath{ {fig/} }
\def\BibTeX{{\rm B\kern-.05em{\sc i\kern-.025em b}\kern-.08em
    T\kern-.1667em\lower.7ex\hbox{E}\kern-.125emX}}

\def\neuimm{{NeuIM}\xspace}

\begin{document}

\title{ Physics-Informed 
Induction Machine Modelling\\}
\author{Qing~Shen,~\IEEEmembership{Student Member,~IEEE}, Yifan~Zhou,~\IEEEmembership{Member,~IEEE}, and Peng~Zhang~\IEEEmembership{Senior Member,~IEEE},
\thanks{
This work was supported in part by the U.S. Department of Energy’s Office of Energy Efficiency and Renewable Energy (EERE)  under the Solar Energy Technologies Office Award Number 38456 
and in part by the National Science Foundation under Grant No. ITE-2134840. 

Q. Shen, Y. Zhou and P. Zhang are with the Department of Electrical and Computer Engineering, Stony Brook University, NY, USA (e-mail: p.zhang@stonybrook.edu).} 
}
 \maketitle
\begin{abstract}
This rapid communication devises a Neural Induction Machine (\neuimm) model, which pilots the use of physics-informed 
machine learning to enable AI-based electromagnetic transient simulations. 
The contributions are threefold: (1) a formation of \neuimm to represent the induction machine in phase domain; (2) a physics-informed 
neural network capable of capturing 
fast and slow IM dynamics 
even in the absence of data; and (3) a data-physics-integrated hybrid \neuimm approach which is adaptive to 
various levels of data availability. 
Extensive case studies validate the efficacy of \neuimm and in particular, its advantage over purely data-driven approaches.
\end{abstract}
\vspace{-6pt}
\section{Introduction}\label{sec:intro}
\IEEEPARstart
{E}{lectromagnetic} Transients Program~\cite{dommel1996emtp} (EMTP), capable of capturing broad-spectrum and high-fidelity dynamics, is indispensable for the planning and operation of today's power systems. Induction  machines (IMs)  form  a  large  portion  of loads, distributed energy resources (DERs) and industrial power systems.  The EMTP simulations of IMs' dynamics 
are critical for the resilient operations of the DERs and associated power grids~\cite{peng}. Small time steps required in the numerical integration of IMs' dynamics, however, make the EMTP solution a formidable problem even on powerful real-time simulators. Ultra-scalable IM models, therefore, are highly needed to circumvent conventional model-based IM modeling and daunting EMTP simulation burdens.   

To bridge the gap, a natural intuition is to integrate the proven ultra-scalability of physics-informed machine learning~\cite{karniadakis2021physics} (PIML) with an accurate physical IM model. Unlike the supervised data-driven learning, which requires a substantial amount of data, unsupervised PIML eliminates the need for extensive data acquisition and storage, and is insensitive to imperfect data and parameter modifications. Among existing IM models, the voltage-behind-reactance  (VBR) model has excellent interpretability and efficiency~\cite{10077403}\cite{4505399}. By mapping all IM state variables into the \emph{abc} stator equations, the VBR model can be directly coupled into the system equations. 
This letter, therefore, fuses PIML and VBR techniques  to devise a physics-informed neural machine  modelling (\neuimm) which not only renders a high-fidelity solution but also takes full advantage of the identified physical mechanism. The compact \neuimm structure eliminates the redundant variables,  preserves the essential physical representations, and can be ultra-efficiently deployed in edge computing devices.

\section{Basics of Physics-Informed Machine Learning }
\label{sec:2}
To give a basic example of physics-informed neural network (PINN), an ordinary differential equation is considered \cite{2019JCoPh.378..686R}: 
\begin{equation}
    \frac{du}{dt}=R(u,\gamma), ~u|_{t=0}=u_0, t \in [0,T]
\end{equation}
where $u(t)$ denotes the state variable, $R$ is a nonlinear operator parametrized by $\gamma$, $T$ is the total time. To solve such an equation, a PINN is defined with the loss function $\mathcal{L}_{phy}$ as: 
\begin{small}
\begin{equation}
\begin{aligned}
    \mathcal{L}_{phy}&=L_{b}+L_p\\
s.t.~~
    L_p&=\frac{1}{n_T}\sum_i^{n_T}||\frac{du_{NN}}{dt}|_{t_i}-R(u_{NN}(t_i),\gamma)||_2\\
     L_{b}&=||u_{NN}(t_0)-u_0||_2,~t_0=0
\end{aligned}   
\end{equation}
\end{small}

\noindent Here $u_{NN}$ is the predicted output from the PINN, $n_T$ is the number of time steps. Instead of directly deploying $\mathcal{L}=\sum_i^{n_T}||u_{NN,i}-u_i||_2$ to calculate the difference between the predictions and real measurements in a data-driven manner,
PINN exploits physical equations to construct the loss $L_p$, and incorporates only the boundary conditions of $u$ to form $L_b$, which collectively guide the training process.
\vspace{-6pt}
\section{Physics-informed \neuimm}
A physics-informed neural network 
is devised to obtain the phase domain 
IM model using sparse 
data. The \neuimm model 
guarantees to follow all the IM physics (a VBR abstraction) 
even when the machine parameters change, which outperforms purely data-driven approaches due to its generalization ability and 
saved efforts of gathering abundant data.\vspace{-10pt}
\subsection{Learning-based model of induction machine}
Leveraging the prior knowledge of the VBR model:
\begin{small}
\begin{equation} \label{equ:electrical:1all}
\begin{aligned}
&\Dot{\lambda}_{qd0s,r}=s(\omega,\omega_r,\lambda_{dq0s,r},r_{s,r},i_{qd0s,r},v_{qd0s})\\
&\lambda_{qd0s,r}=f(L_{ls,lr},L_M,i_{qd0s,r})
\end{aligned}
\end{equation}
\end{small}
where $f$, $s$ denote the physics laws of the flux linkages $\lambda_{qd0s,r}$~\cite{peng}. $L_{ls,lr},r_{s,r},L_M$ are machine parameters. The learning-based model of the IM can then be formulated as:
\begin{small}
 \begin{equation} \label{equ:NN}
     \frac{di_{abcs}}{dt}=N(v_{abcs},z), \qquad i_{abcs}|_{t=0}=i_0, i_{qd0s}|_{t=0}=\tilde{i}_0
 \end{equation}  
\end{small}

where $v_{abcs}$ is the terminal
voltages on the stator side, $N$ is a neural network to predict the derivatives of $i_{abcs}$. $i_{0}$, $\tilde{i}_0$ are the initial values of $i_{abcs}$, $i_{qd0s}$. 
 \vspace{-8pt}
\subsection{Physics-informed learning for \neuimm}
The kernel idea is to take advantage of the well-established physics laws of IM to assist the training of \neuimm. $N$ is divided into two sub-neural models, $G$ and $P$ for two reasons. Firstly, performing the Park inversion in training results in slow computation speed due to the need for calculating it at each time step using the dynamic variable $\theta$. Secondly, separating the Park inversion from the training process enables a more direct and informative loss function, allowing for a clearer understanding of each neural network's behavior.

\begin{figure}[!ht]
\vspace{-12pt}
    \centering
    \includegraphics[width=1\columnwidth]{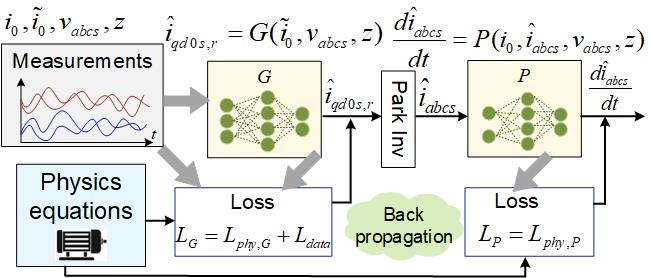}
    \caption{Schematic diagram of physics-informed \neuimm}
    \vspace{-10pt} 
    \label{fig:sys1}  
\end{figure}

Here \textsuperscript{$\wedge$} denotes the predicted values. As shown in Fig.~\ref{fig:sys1}, with the measurements of $[\theta,\omega,\omega_r]$ as the input $z$, $v_{abcs}$ and initial values, $G$ will yield $\hat{i}_{qd0s,r}$. Then $\hat{i}_{abcs}$ is obtained via Park's inverse transformation and passed into $P$ to get the final output  $\hat{i}_{abcs}/dt$.

$\lambda$ is readily computed using the outputs of $G$~\cite{peng}. 
Further, $\lambda'$ is defined according to modified Euler to enhance estimation precision and guarantee the uniqueness of solution:
\begin{equation} \label{equ:loss:1}
\begin{aligned}
&\lambda(t)=f(G,L_{ls,lr},L_M), ~
\dot{\lambda}(t)=s(\lambda(t),z,r_{s,r},G)\\
  & \lambda'(t+1) = \lambda(t) + dt \cdot \dot{\lambda}(t)\\
   &\Dot{\lambda'}(t+1)  =s(\lambda'(t+1),z,r_{s,r},G)
\end{aligned}
\end{equation}
\noindent 
Finally, the construction of loss for $G$ 
is developed as:
\vspace{-5pt}
\begin{small}
\begin{equation} \label{equ:phy loss}
\begin{aligned} 
 \min_{{\theta}} L_{G}& = L_{phy,G}
= \sum_{i=1}^m \sum_{t=1}^{T} \| \delta_i(t) - \delta'_i(t) \|_2\\
s.t.~~ \delta_i(t)&=\lambda_i(t+1) -\lambda_i(t), \\ \delta'_i(t)&=\frac{dt}{2} \cdot [\dot{\lambda'_{i}}(t+1)+\dot{\lambda_i}(t)]
\end{aligned}
\end{equation}
\end{small}
\noindent $m$ is 
the number of selected intermediate variables. For instance, 
here $m=6$ representing the six $\lambda$s. Similarly, the training model for $P$ is developed as:
\vspace{-5pt}
\begin{small}
\begin{equation}
\begin{aligned}
\min_{{\theta}}L_{P}& = \sum_{j=1}^l \sum_{t=1}^{T} \| \Delta_j(t) - \Delta'_j(t) \|_2 \\
 s.t.~~\Delta_j(t)&=\hat{i}_{abcs,j}(t+1) -\hat{i}_{abcs,j}(t), \\ \Delta'_j(t)&=\frac{dt}{2} \cdot [\frac{{d\hat{i}}'_{abcs,j}(t+1)}{dt}+\frac{d\hat{i}_{abcs,j}(t)}{dt}]
 \end{aligned}
\end{equation}
\end{small}
where $ \hat{i}_{abcs}$ is derived from $G$, the second network $P$ outputs $d{\hat{i}_{abcs}}/{dt}$, $l=3$ represents the three phases $a,b,c$. 
\vspace{-6pt}
\subsection{Data and physics hybrid \neuimm}
One complication of the purely physics-informed (PI) method is that it requires long time to get converged, especially for fast transient cases. Interestingly, if limited data is available, the efficiency of training is found extensively improved. Therefore, a data-physics hybrid \neuimm is proposed to unlock the potential \cite{9743327}. A hybrid \neuimm can be regarded as a grey box with a compelling advantage that, provided only limited data to equip with the training, it can achieve a generalization ability which the purely data-driven approach may require far more data to reach; whereas, data-driven NN can not converge if the data is not complete. Hybrid \neuimm thus ignites the hope to make up for the incomplete data set by deploying physics information. 

For hybrid \neuimm, only the construction of loss in $G$ is different:
\begin{small}
\begin{equation}
\begin{aligned}
 \min_{{\theta}} L_{G}&= L_{phy,G}+ L_{data} 
\\
s.t.~~ 
 L_{data}& =\sum_{k=1}^n\sum_{t=1}^{T}  \| \hat{i}_{abcs,k}(t) - i_{abcs,k}(t) \|_2 
\end{aligned}
\end{equation}
\end{small}
where $L_{phy,G}$ stands for the purely physical loss defined in \eqref{equ:phy loss} and $L_{data}$ is the loss calculated from ground truth $i_{abcs}$. $n$ is the number of available trajectory with complete data. If $n=0$, then the construction of loss is purely physics-informed; otherwise, the loss is a hybrid of physics and data. \neuimm's algorithm is listed below. $Threshold_G$ and $Threshold_P$ are adjustable to expected accuracy.
\SetKwInput{KwInput}{Inputs}                
\SetKwInput{KwResults}{Results}
\begin{algorithm}
\begin{small}
  \caption{\neuimm }\label{alg:two}
\KwInput{$i_0,\tilde{i}_0,v_{abcs},z$}
\While{$t < T$}{
  \eIf{$L_G >  threshold_G$ }{
    $\hat{i}_{qd0s,r} \gets G(i_0,v_{abcs},z)$
    
    update $L_G{=}L_{phy} +L_{data}$ (hybrid \neuimm), back propagate
  } 
  {$\hat{i}_{abcs} \gets ParkInv(\hat{i}_{qd0s},\theta)$
  \\\If{$L_P >  threshold_P$}{
      $d\hat{i}_{abcs}/{dt}\gets P(\hat{i}_{abcs},v_{abcs},z)$, update $L_P$ then back propagate
    }
  }
}  
\KwResults{\neuimm model as depicted in Eq.~\eqref{equ:NN}}
\end{small}
\end{algorithm}
\vspace{-8pt}
\section{Case Study}
This section 
verifies the excellent performance of \neuimm under various operational conditions 
and demonstrate the advantages of the hybrid \neuimm over the data-driven method.  \neuimm is 
deployed with Tensorflow 1.5 (Python 3.6) 
with 2 hidden layers. The ground truth of the IM dynamics is obtained by running the original VBR model in Matlab. The free acceleration and torque change case use a 3-hp machine, 2500-hp machine is used for the fault case. Detailed test data are given in~\cite{peng}, and key machine parameters are listed below:
\vspace{-14pt}
\begin{table}[ht]
\begin{center}
  \caption{Machine parameters}\label{table1} \vspace{-6pt}
\begin{tabular}{ccccc}
    \hline
      Type &$r_{s,r}$/ $\Omega$ &$L_{ls,lr}$/ $\Omega$&$L_M$/ H &line-line voltage \\
       \hline
    3-hp &0.435;0.816 &0.754;0.754& 0.0693&220V\\ 
     2500-hp&0.029;0.022&0.226;0.226 &0.0346&2.3Kv\\ 
    \hline
\end{tabular}
\end{center}
\end{table}
\vspace{-26pt}
\subsection{Efficacy of \neuimm under varied operational conditions}
The efficacy of \neuimm is validated in three scenarios, free acceleration, torque changes, faults, as specified in Table~\ref{table1}:
\vspace{-6pt}
\begin{table}[ht]
\begin{center}
  \caption{Data sets}\label{table1} \vspace{-6pt}
\begin{tabular}{ccccc}
    \hline
      Scenario &Data set &torque/(N$\cdot $m)&Voltage/kV&$L_M/H$ \\
       \hline
    \multirow{2}{*}{Torque change (3-hp)}&training &$\pm$5,10 &0.22&0.0693\\ &testing&$\pm$ 3,12&0.22&0.0693 \\
     \multirow{2}{*}{Fault (2500-hp)}&training&8900 &2.3, 2.4,2.5&0.0346\\ &testing&8900&2.3,2.35&0.0531\\
    \hline
\end{tabular}
\end{center}
\end{table}
\vspace{-12pt}

In the load torque change tests, for instance, 
the mechanical torque $T_m$ changes from 0 to 12 N$\cdot$m at 2.05s and stays at 12 N$\cdot$m until 2.5s, then $T_m$ is reversed 
to $-$12 N$\cdot$m, stays till the end. This case is marked $\pm$12 in Table~\ref{table1}. As an example of the fault cases, 
a three-phase fault is applied at 6.1s and cleared at 6.2s. Moreover, the parameter $L_M$ 
changes in the testing of fault cases to show the adaptability of NeuIM.

\begin{figure}[!ht]
\vspace{-4pt}
      \includegraphics[width=0.6\columnwidth]{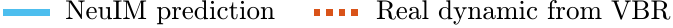}
     \begin{subfigure}{0.47\columnwidth}
        \includegraphics[width=\columnwidth]{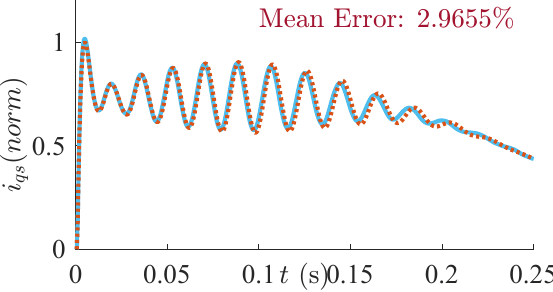}
        \caption{prediction of G under free acceleration}
         \label{fig:i_acc}
    \end{subfigure}     
    \begin{subfigure}{0.47\columnwidth}
        \includegraphics[width=\columnwidth]{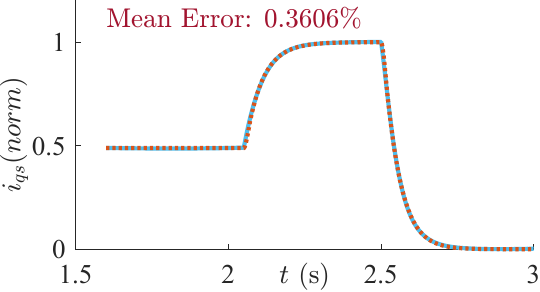}
        \caption{prediction of G under torque change}
        \label{fig:i_tor}
    \end{subfigure}
     \begin{subfigure}{0.47\columnwidth}
        \includegraphics[width=\columnwidth]{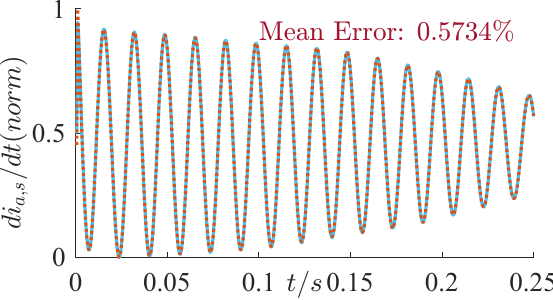}
        \caption{prediction of P under free acceleration}
        \label{fig:di_acc}
    \end{subfigure}
     \begin{subfigure}{0.47\columnwidth}
        \includegraphics[width=\columnwidth]{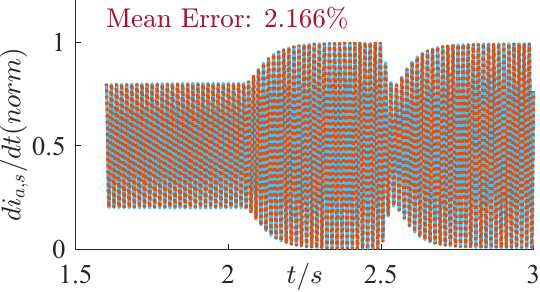}
        \caption{prediction of P under torque change}
        \label{fig:di_tor}
    \end{subfigure}
    \vspace{-3pt}
    \caption{Performance of \neuimm under free acceleration and torque change}
\vspace{-10pt}
\label{fig:acc&tor}
\end{figure}

Fig.~\ref{fig:acc&tor} presents the performance of \neuimm under free acceleration and torque changes. Fig.~\ref{fig:i_acc}-\ref{fig:i_tor} show the accuracy of predictions from the first neural network $G$; Fig.~\ref{fig:di_acc}-\ref{fig:di_tor} show the final results of $P$. Trajectories of predicted current $i_{q,s}$ and the final output of $di_{a,s}/{dt}$ demonstrate a perfect match between \neuimm's results and real dynamics, verifying the accuracy of 
\neuimm in capturing the relatively slow dynamics. 
\vspace{-10pt}

\begin{figure}[!ht]
\vspace{-6pt}
    \centering
      \includegraphics[width=\columnwidth]{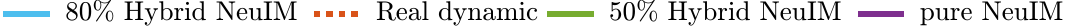}
     \begin{subfigure}{0.47\columnwidth}
        \includegraphics[width=\columnwidth]{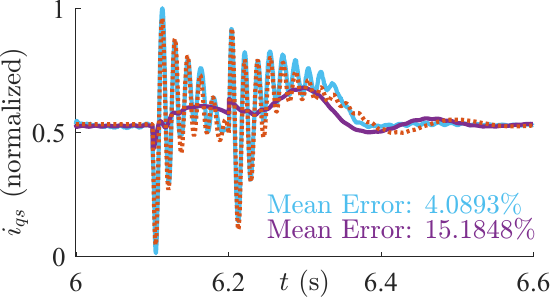}
        \caption{Varied hybrid NeuIM performance}
         \label{fig:hybrid}
    \end{subfigure}
     \begin{subfigure}{0.47\columnwidth}
        \includegraphics[width=\columnwidth]{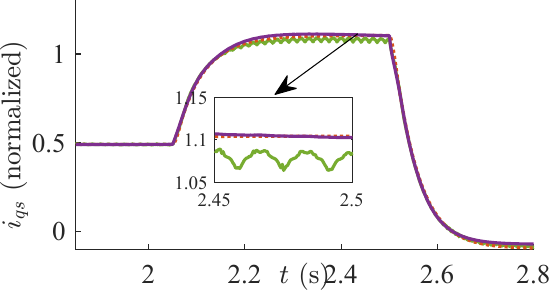}
        \caption{purely PI \neuimm comparison}
        \label{fig:data vs neu2}
    \end{subfigure}   
     \begin{subfigure}{0.9\columnwidth}
    \includegraphics[width=\columnwidth]{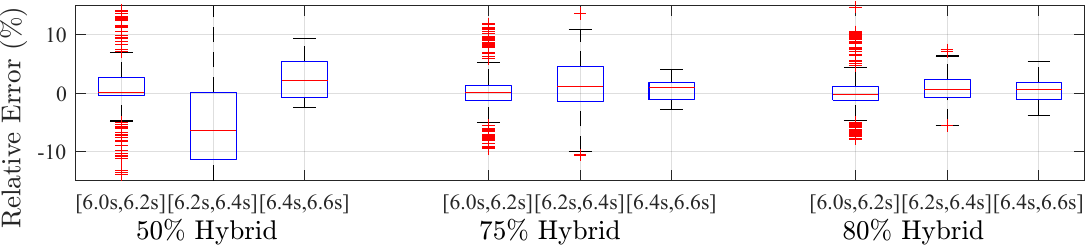}
    \caption{Relative error comparions among varied hybrid \neuimm}
    \label{fig:hy1}
    \end{subfigure}
     \caption{Performance and comparions of \neuimm } \label{fig:vs}
\vspace{-8pt}
\end{figure}

 As for the first fault case, the voltage starts at 2.3 kV; then at 6.01s, a three phase short circuit is applied at the terminals; later at 6.11s, the fault is cleared. In all the training set, the parameter $L_M$  
is 0.0346 H while in the testing set of the fault case, $L_M$ is 0.0531 H.
Fig.~\ref{fig:hy1} illustrates the performance of the hybrid \neuimm under different portions of data. For instance, 75\% hybrid means that 75\% of training trajectories have the true values of the outputs of $G$, i.e. $i_{qd0s,r}$; thus the loss of these trajectories is $L_{phy}+L_{data}$. For the other 25\% of the trajectories where the true values of $i_{qd0s,r}$ are unavailable, the loss of these remaining subset only consists of $L_{phy}$. The training philosophy for $P$ is always purely physics-informed.   

 From Fig.~\ref{fig:hy1}, when the proportion of data is 50\%, the performance is not satisfiable. When the proportion is close to 100\%, the data-driven loss takes higher weight in the total loss function. Even though the training accuracy is higher, the testing accuracy and generalization ability are compromised if it is purely data-driven. Therefore, in the next section, the optimal percentage of data, 80\% data is used. Fig.~\ref{fig:hybrid} shows the 80\% \neuimm performances beats pure \neuimm.

\vspace{-8pt}
\subsection{Comparison with data-driven approach}
  In Fig.~\ref{fig:data vs neu2}, it is observable that the purely data-driven deep neural network (DNN), which has the same structure as \neuimm, has irregular oscillations due to its training with limited data. Meanwhile, \neuimm is not impacted by this issue. 
\vspace{-12pt}
\begin{table}[ht]
\begin{center}
\caption{MSE comparsion}\label{tab:mse}
\vspace{-4pt}
\begin{tabular}{ccc}
    \hline
     Type &Fault&Torque change\\
     \hline
      Purely data-driven&0.0386&0.4330\\
      Purely PI \neuimm&0.7352&0.0182\\
      80\% Hybrid \neuimm&0.0142& 0.0158\\
    \hline
\end{tabular}
\end{center}
\end{table}
\vspace{-12pt}

Table~\ref{tab:mse} shows the final results from $P$. Hybrid physics-informed \neuimm overshadows the purely data-driven DNN in terms of accuracy, which reveals that under the same training set, \neuimm has better generalization ability to cope with unseen cases and varied parameters. For slow transients such as torque change, both \neuimm and hybrid \neuimm beat the data-driven approach. For the fault cases, the hybrid \neuimm outperforms both the data-driven method and the purely physical-informed \neuimm in terms of better convergence rates and lower MSE. The data-driven approach, even though it outperforms the pure \neuimm when data is available, it cannot grasp the transients completely when one machine parameter changed. In addition, as in Table~\ref{tab:time}, by checking the overall calculation time, \neuimm is significantly more efficient and its scalablity is guaranteed.
\vspace{-3pt}
\begin{table}[ht]
\begin{center}
  \caption{Computation time of VBR ($\Delta t=0.5ms$) and \neuimm}\label{tab:time}
  \vspace{-4pt}
\begin{tabular}{ccccc}
    \hline
     Scenario& \neuimm  &VBR   \\
       \hline
        Free Acceleration ($T=1s$) &0.1684$s$&0.4756$s$\\
        Torque change ($T=3.5s$) &0.5514$s$& 1.2751$s$\\
   Fault ($T=10s$)&0.5553$s$ (80\% hybrid)&2.0727$s$\\
    \hline
\end{tabular}
\end{center}
\vspace{-17pt}
\end{table}
\section{Conclusion}
This communication devises physics-informed Neural Induction Machine Modelling (\neuimm) and its variant  hybrid \neuimm. 
Case studies for \neuimm and hybrid \neuimm show the efficacy of the 
new methods under various
contingencies, and their superiority over existing purely data-driven methods. The success of NeuIM is the first step to develop an AI-driven grid simulator that replaces conventional model-based simulators. 
\bibliographystyle{ieeetr}
\bibliography{ref}

\vspace{12pt}

\end{document}